\documentclass[runningheads]{llncs}
\usepackage{graphicx}

\usepackage{tikz}
\usepackage{comment}
\usepackage{amsmath,amssymb} 
\usepackage{color}
\usepackage{orcidlink}

\usepackage[accsupp]{axessibility}  

\usepackage{algorithm}
\usepackage{algpseudocode}

\usepackage{array}
\usepackage{subcaption}
\newcolumntype{C}[1]{>{\centering\arraybackslash}p{#1}}

\begin{document}
\pagestyle{headings}
\mainmatter
\def\ECCVSubNumber{893}  

\title{
LiDAL: Inter-frame Uncertainty Based Active Learning for 3D LiDAR Semantic Segmentation
} 

\titlerunning{LiDAL}

\author{Zeyu Hu\inst{1}\orcidlink{0000-0003-3585-7381}\thanks{intern at Tencent Lightspeed Studios} \and 
 Xuyang Bai\inst{1}\orcidlink{0000-0002-7414-0319} \and
 Runze Zhang\inst{2}\orcidlink{0000-0001-9698-0178} \and
 Xin Wang\inst{2}\orcidlink{0000-0002-6789-4569} \and
 Guangyuan Sun\inst{2}\orcidlink{0000-0002-5595-0594} \and
 Hongbo Fu\inst{3}\orcidlink{0000-0002-0284-726X} \and
 Chiew-Lan Tai\inst{1}\orcidlink{0000-0002-1486-1974}
}

\authorrunning{Z. Hu et al.}

\institute{Hong Kong University of Science and Technology\\
\email{\{zhuam,xbaiad,taicl\}@cse.ust.hk}\\
\and
Lightspeed \& Quantum Studios, Tencent\\
\email{\{ryanrzzhang,alexinwang,gerrysun\}@tencent.com}
\and
City University of Hong Kong\\
\email{hongbofu@cityu.edu.hk}
}

\maketitle

\begin{abstract}
We propose LiDAL, a novel active learning method for 3D LiDAR semantic segmentation {by exploiting} 
inter-frame uncertainty among LiDAR {frames}. 
{Our core idea is that a well-trained model should generate robust results irrespective of {viewpoints for scene scanning} 
and thus the inconsistencies in model predictions across frames 
provide a very reliable
measure of uncertainty for active sample selection.}
To implement this uncertainty measure, we introduce new inter-frame divergence and entropy formulations, which serve as the metrics for active selection. Moreover, we demonstrate additional performance gains by {predicting and incorporating} 
pseudo-labels, which are also selected using the proposed inter-frame uncertainty measure.
Experimental results validate the effectiveness of LiDAL: we achieve 95\% of the performance of fully supervised learning with less than 5\% of annotations on the SemanticKITTI and 
nuScenes datasets, outperforming state-of-the-art active learning methods. Code release: \url{https://github.com/hzykent/LiDAL}
\end{abstract}

\section{Introduction}
Light detection and ranging (LiDAR) sensors {capture} 
more precise and farther-away distance measurements than conventional visual cameras, {and} have become a necessity for an accurate perception system of outdoor scenes. 
{These sensors} 
generate 
rich 3D geometry {of real-world scenes as 3D point clouds} 
to facilitate a thorough scene understanding, in which 3D LiDAR semantic segmentation serves as a cornerstone.
{The {semantic segmentation} task is to parse a scene and assign}
{an object} 
class label to each point in 3D point clouds, {thus} providing point-wise perception information for numerous downstream applications like robotics~\cite{thrun2006stanley} and autonomous vehicles~\cite{li2016vehicle}.

\begin{figure}[htp]
\centering
\includegraphics[width=\textwidth]{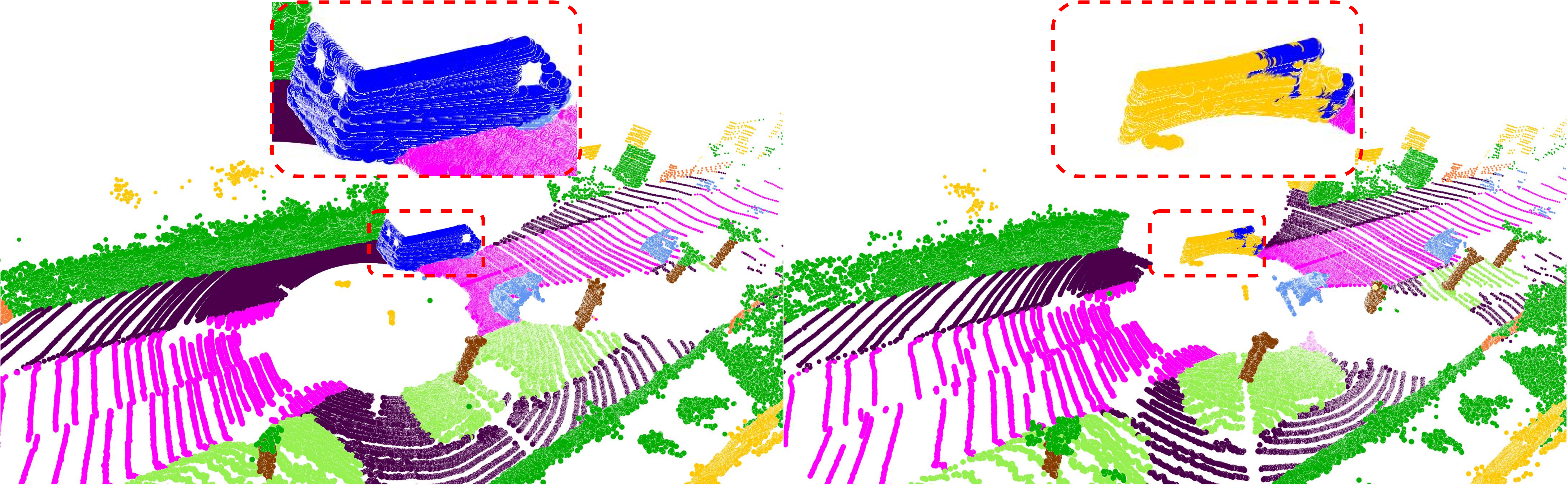}
\caption{\textbf{Illustration of inter-frame uncertainty.} While in {one} 
frame {(Left) an} 
object is correctly predicted as ``vehicle'' {(highlighted in a red box)}, in the {subsequent} 
frame {(Right)} a large part of {this object} 
is mistakenly predicted as ``fence'' when scanned from a different viewpoint. }
\label{fig:motivation}
\end{figure}

Thanks to the large-scale LiDAR datasets~\cite{behley2019semantickitti,caesar2020nuscenes} made public{ly} available in recent years,
the state of the art in 3D LiDAR semantic segmentation has been significantly pushed forward~\cite{thomas2019kpconv,hu2020randla,zhu2021cylindrical}. However, the requirement of fully labeled point clouds for existing {segmentation} methods has become a major obstacle to scaling up the {perception} system or extending it to {new scenarios.} 
Typically, {since} a LiDAR sensor may perceive millions of points per second, exhaustively labeling all points is extremely laborious and time-consuming. It poses demands on developing label-efficient approaches for 3D LiDAR semantic segmentation.

Active learning provides a promising solution to reduce the costs associated with labeling. {Its} 
core idea is to design a learning algorithm that can interactively query a user to label new data samples according to a {certain} policy, leading to models trained with only a fraction of the data while yielding similar performances. Inspired by 2D counterparts~\cite{vezhnevets2012active,gorriz2017cost,mackowiak2018cereals,li2020attention,siddiqui2020viewal}, some previous works have explored active learning for 3D LiDAR semantic segmentation~\cite{luo2018semantic,lin2020active,wu2021redal,shi2021label}. However, {these} 
methods almost exclusively operate on single LiDAR frame{s. Such a strategy} 
is surprising since,
{unlike most 2D datasets, in which images are captured as independent samples, 3D LiDAR datasets are generally scanned as continuous point cloud frames.}
As a consequence, inter-frame constraints naturally embedded in the LiDAR scene sequences are largely ignored{. We believe such constraints} 
are particularly interesting for examining the quality of network predictions; i.e., the same object in a LiDAR scene should receive the same label when scanned from different viewpoints (see Fig.~\ref{fig:motivation}).

In this work, we propose to exploit inter-frame constraints in a novel view-consistency-based uncertainty measure. More specifically, we propose new inter-frame divergence and entropy formulations based on the variance of predicted score functions across continuous LiDAR frames. For a given (unlabeled) object (e.g., {the one in a red box} 
in Fig.~\ref{fig:motivation}), if its predicted labels differ across frames, we assume faulty network predictions and then strive to obtain 
{user-specified}
labels for the most uncertain regions. In addition to the main active learning formulation, we also explore further improvements to the labeling efficiency with self-training by utilizing the proposed uncertainty measure in an inverse way. During each active learning iteration, we augment the 
{user-specified}
labels with pseudo-labels generated from the most certain regions across frames to further boost performance without extra annotations or much computational cost. 

To summarize, our contributions are threefold: 
    \begin{enumerate}
        \item We propose a novel active learning strategy for 3D LiDAR semantic segmentation {by estimating} 
        model uncertainty based on the inconsistency of predictions across frames.
        \item We explore self-training in the proposed active learning framework and show that further gains can be realized by including pseudo-labels.
        \item Through extensive experiments, we show that the proposed active learning strategy and self-training technique significantly improve labeling efficiency over baselines, and establish the state of the art in active learning for 3D LiDAR semantic segmentation. 
    \end{enumerate}

\section{Related Work}
Compared to fully supervised methods~\cite{thomas2019kpconv,hu2020jsenet,zhu2021cylindrical,tang2020searching,choy20194d,hu2021vmnet}, label-efficient 3D semantic segmentation is a relatively open research problem. Previous explorations can be roughly divided into five categories: transfer learning, unsupervised and self-supervised learning, weakly-supervised learning, active learning, and self-training. LiDAL falls into both the active learning and self-training categories.

\textbf{Transfer Learning.} 
Taking advantage of existing fully labeled datasets, transfer learning has been introduced to 3D semantic segmentation for reducing the annotation costs. Various domain adaptation approaches have been developed to make {them} 
perform well in novel scenarios given only labeled data from other domains~\cite{liu2021adversarial,yi2021complete,langer2020domain} or even synthetic training sets~\cite{wu2019squeezesegv2}. They achieve fairly decent results but still require fully labeled data from {a} 
source domain 
and fail to generalize to {new} scenarios that are highly different from the source.

\textbf{Unsupervised and Self-supervised Learning.}  
Leveraging the colossal amount of unlabeled data, pre-trained models 
{can be fine-tuned on a small set of labeled data to alleviate the over-dependence on labels and thus achieve satisfactory performances~\cite{hassani2019unsupervised,thabet2019mortonnet,sun2020canonical,xu2021image2point,huang2021spatio}.} 
Pseudo tasks used for pre-training include reconstructing space~\cite{sauder2019self}, contrast learning~\cite{xie2020pointcontrast,liu2020p4contrast,hou2021exploring}, ball cover prediction~\cite{sharma2020self}, instance discrimination~\cite{zhang2021self}, and point completion~\cite{wang2021unsupervised}, etc. Compared to other label-efficient counterparts, these methods require more labeled data and most of them only apply to object-level point clouds.

\textbf{Weakly-supervised Learning.}
Instead of point-by-point labeling in fully supervised learning, weak labels take various forms like scene-level or sub-cloud-level label{s}~\cite{ren20213d,wei2020multi}, 2D supervision~\cite{wang2020weakly}, fewer point label{s}~\cite{xu2020weakly,cheng2021sspc,zhang2021weakly,hu2021sqn,wei2021dense}, seg-level label{s}~\cite{liu2021one,tao2020seggroup}, and box-level label{s}~\cite{liu2022box2seg}, etc. These methods can reduce the number of labeled samples, but either require intricate labeling processes or produce much {more} inferior results than the fully-supervised counterparts.

\textbf{Active Learning.}
During network training, active learning methods iteratively select the most valuable data for label acquisition. The very few existing methods {have explored} 
uncertainty measurements like segment entropy~\cite{lin2020efficient}, color discontinuity~\cite{wu2021redal}, and structural complexity~\cite{wu2021redal,shi2021label}. {The existing} 
methods take LiDAR {data} 
as separated frames and only {consider} intra-frame information. 
Inspired by a 2D work operating on multi-view images~\cite{siddiqui2020viewal}, we take advantage of the inter-frame constraints for active learning in this work. {Different from this 2D work, due to the distinct natures of 2D images and 3D point clouds, we design novel uncertainty formulations and selection strategies. Moreover, we propose a joint active learning and self-training framework to further exploit the inter-frame constraints.}

\textbf{Self-training.}
Building on the principle of knowledge distillation, previous methods generate pseudo labels to expand 
sparse labels~\cite{xu2019semantic,liu2021one} or to facilitate the network training using only scene-level supervision~\cite{ren20213d}. In this
work, we develop a pseudo-labeling method applied in conjunction with our active learning framework {to achieve} 
even greater gains in efficiency.

\section{Method}

\begin{figure}[t]
\centering
\includegraphics[width=\textwidth]{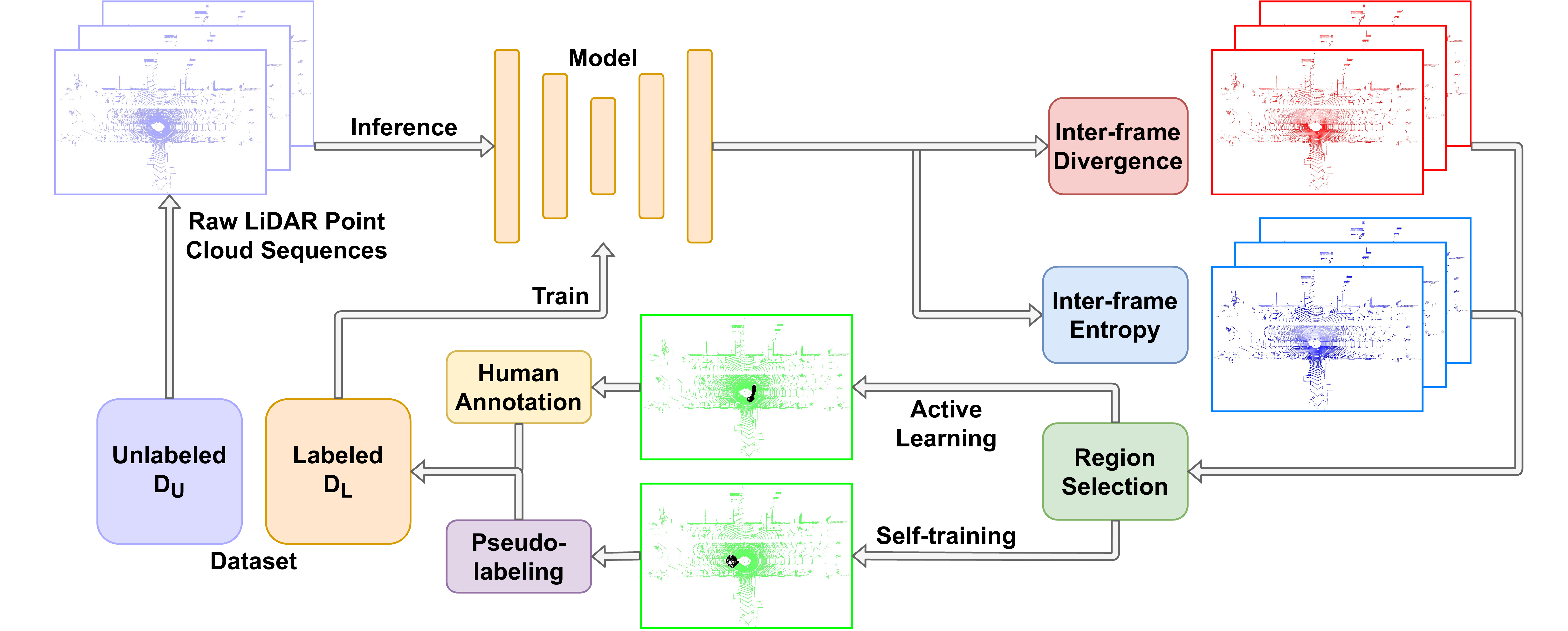}
\caption{\textbf{Pipeline of LiDAL.} In each round of active learning, we first train a 3D LiDAR semantic segmentation network in supervision with labeled dataset $D_L$. Second, we use the trained network to compute an inter-frame divergence {score} and an inter-frame entropy score for all regions from the unlabeled dataset $D_U$. We then select a batch of regions based on these scores for active learning and self-training, and finally request their respective labels from the human annotation and the pseudo-labeling. The process is repeated until the labeling budget is exhausted or all training data is labeled.}
\label{fig:pipeline}
\end{figure}

\subsection{Overview}\label{sec:overview}

The goal of LiDAL is to train a well-performing 3D LiDAR semantic segmentation model with a constrained annotation budget. Specifically, we assume {the} availability of data $D = \{D_L, D_U\}$. {The data consists of sequences of LiDAR frames, each provided with an ego-pose of the scanning device.} Initially, $D_L$ is a small {set of LiDAR frames randomly selected
from $D$ with each frame having its label annotation}, 
and $D_U$ is a large unlabeled set without any annotations. Following previous works~\cite{siddiqui2020viewal,wu2021redal}, we use a sub-scene region as {a} 
fundamental query unit to focus on the most informative parts of the LiDAR frames.

{As illustrated in Fig.~\ref{fig:pipeline}}, our LiDAL method consists of four main steps: 
1. Train the network to convergence using the currently labeled dataset $D_L$. 2. Calculate the model uncertainty scores for each region of $D_U$ with two indicators: inter-frame divergence and inter-frame entropy (Section~\ref{sec:score}). 3. Select regions based on {the} uncertainty measures for active learning and self-training (Section~\ref{sec:region}). 4. Obtain labels from human annotation and pseudo-labeling. {These steps will be repeated until the labeling budget is exhausted {or all training data is labeled}.}

\subsection{Uncertainty Scoring}\label{sec:score}
\begin{figure}[t]
\centering
\includegraphics[width=\textwidth]{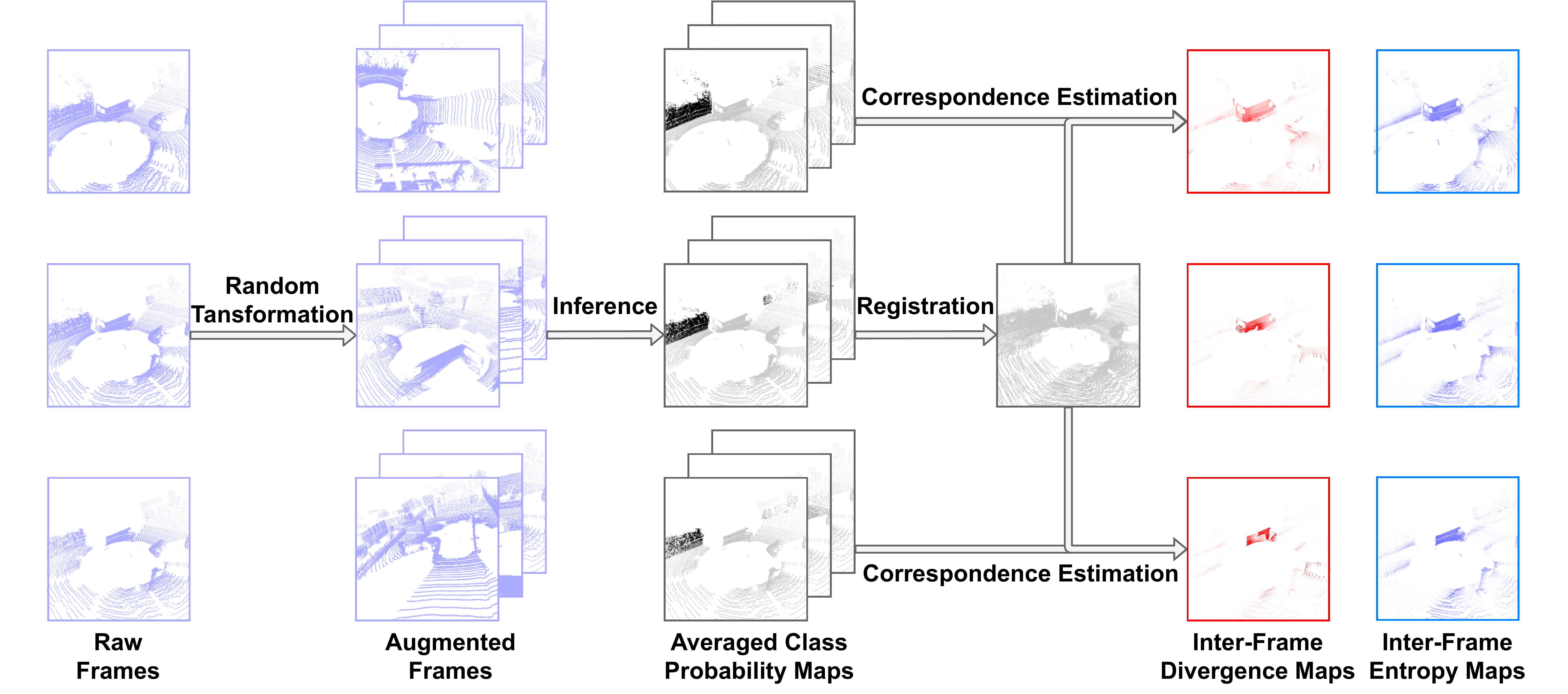}
\caption{\textbf{Illustration of uncertainty scoring.} For each unlabeled LiDAR frame in the dataset, we first obtain its averaged class probability predictions from augmented inference runs. Next, we register each frame with its provided ego-pose. For each point in the frame, we then find its corresponding points in neighboring frames and assign to it their associated class probability predictions. With the aggregated multiple class probability predictions per point, we compute the inter-frame divergence and entropy scores. We assign these two scores to each unlabeled region by averaging the scores of all the points contained in it. 
}
\label{fig:uncertainty}
\end{figure}

At each iteration,
{after the first step of training the network on $D_L$,}
our active learning method LiDAL {then}
aims at predicting which samples from $D_U$ are the most informative to the network at the current state. To this end, we introduce two novel uncertainty scoring metrics named \emph{inter-frame divergence} and \emph{inter-frame entropy}. Fig.~\ref{fig:uncertainty} provides an overview of the scoring process.

\subsubsection{Inter-frame Divergence.}
In a nutshell, the proposed inter-frame divergence score aims at estimating which objects are consistently predicted the same way, irrespective of the scanning viewpoints. 

For each frame, we first calculate its point-wise class probability maps using the current trained segmentation network. 
{To attain robust probability predictions,}
we perform data augmentations with random translation, rotation, scaling, and jittering. The probability $P$ for a point $p$ in frame $F_i$ to belong to class $c$ is given by:
    \begin{equation}
    P_i^p(c) = \frac{1}{D}\sum_{d=1}^DP_{i,d}^p(c),
    \end{equation}
where $D$ is the number of augmented inference runs of the segmentation network, and $P_{i,d}^p(c)$ is the softmax probability of point $p$ belonging to class $c$ in the augmented inference run $d$.

Next, using the provided ego-pose, we register each frame in the world coordinate system. For each point in a given frame, we find its corresponding points in the neighboring frames
and assign to it their associated probability distributions.
{Implementation details can be found in \textbf{Supplementary Section A}.} Each point $p$ in frame $F_i$ is now associated with a set of probability distributions $\Omega_i^p$, each coming from a {neighboring} 
frame:

    \begin{equation}
    \Omega_i^p = \{P_j^p, j  | F_j^p \text{ corresponds to } F_i^p\},
    \end{equation}
{where $F_j^p$
denotes the point $p$ in frame $F_j$, and $F_j$ represents one of the neighboring frames of $F_i$.}

{In our setting, when estimating the point correspondences between neighboring frames, we assume that the objects in the scene are static and thus {the} points in the same registered position represent the same object. The moving objects are not specially treated for two reasons.}
First, they contribute only a small portion of the dataset. Second, when estimating correspondences after registration, the prediction disagreements introduced by the 3D motions can be seen as inter-frame inconsistency and help the system select these informative regions.  

The inter-frame divergence corresponds to the average pairwise KL divergence between the class distribution at any given point and the class distributions assigned to that point {from the neighboring frames}. {It effectively captures} 
the {degree} 
of agreement between the prediction in the current frame with the prediction{s} coming from the neighboring frames. {Specifically, we define} the inter-frame divergence score ${FD}$ for a point $p$ in frame $F_i$ {as follows}: 
    \begin{equation}\label{equ:divergence}
    {FD}_i^p = \frac{1}{|\Omega_i^p|}\sum_{P_j^p\in\Omega_i^p}D_{KL}(P_i^p||P_j^p),
    \end{equation}
where $D_{KL}(P_i^p||P_j^p)$ is the KL Divergence between distributions $P_i^p$ and $P_j^p$.

\subsubsection{Inter-frame Entropy.} 


After measuring how inconsistent the predictions are across frames, we then define the inter-frame entropy score{, which indicates} 
the amount of uncertainty for the network to process a certain point. For a point $p$ in frame $F_i$, with the aggregated probability distributions $\Omega_i^p$, the mean distribution $M_i^p$ can be calculated as:
    \begin{equation}
    M_i^p = \frac{1}{|\Omega_i^p|}\sum_{P_j^p\in\Omega_i^p}P_j^p,
    \end{equation}
which can be seen as {the marginalization of}
the prediction probabilities over the scanning viewpoints.

The inter-frame entropy score ${FE}$ is defined as the entropy of the mean class probability distribution $M_i^p$:
    \begin{equation}\label{equ:entropy}
    {FE}_i^p = -\sum_{c}M_i^p(c)\log (M_i^p(c)).
    \end{equation}

A high inter-frame entropy score implies that on average, the prediction of the current network for this point is significantly uncertain. Since the mean class probability distribution is the average result from both the augmented inference runs and the aggregation of corresponding points, the inter-frame entropy score estimates both the intra-frame uncertainty under random affine transformations and {inter-frame uncertainty under viewpoint changes.} 

\subsection{Region Selection}\label{sec:region}
To select the most informative parts of the unlabeled dataset, we opt for using sub-scene regions as the fundamental label querying units, following previous works~\cite{siddiqui2020viewal,wu2021redal}. Our implementation uses the constrained K-means clustering~\cite{bradley2000constrained} algorithm for region division. An ideal sub-scene region consists of one or several object classes and is lightweight to label for the annotator.

For each region $r$, the two scores ${FD}_i^r$ and ${FE}_i^r$ are computed as the average of the inter-frame divergence and inter-frame entropy scores of all the points contained in $r$:
    \begin{equation}
    {FD}_i^r = \frac{1}{|r|}\sum_{p \in r}{FD}_i^p,
    \end{equation}
    \begin{equation}
    {FE}_i^r = \frac{1}{|r|}\sum_{p \in r}{FE}_i^p,
    \end{equation}
where $|r|$ is the number of points contained in region $r$.

\subsubsection{Active Learning.}
We now discuss our active learning strategy utilizing the proposed inter-frame uncertainty scores. Our strategy to select the next region for labeling consists of two steps. First, we look for the region $r$ from frame $F_i$ that has the highest inter-frame divergence score in $D_U$:
    \begin{equation}
    (i, r) = \mathop{\arg\max}_{(j, s) \in D_U}{FD}_j^s,
    \end{equation}
where $(j, s)$ refers to region $s$ from frame $F_j$.

\begin{figure}[t]
\centering
\includegraphics[width=\textwidth]{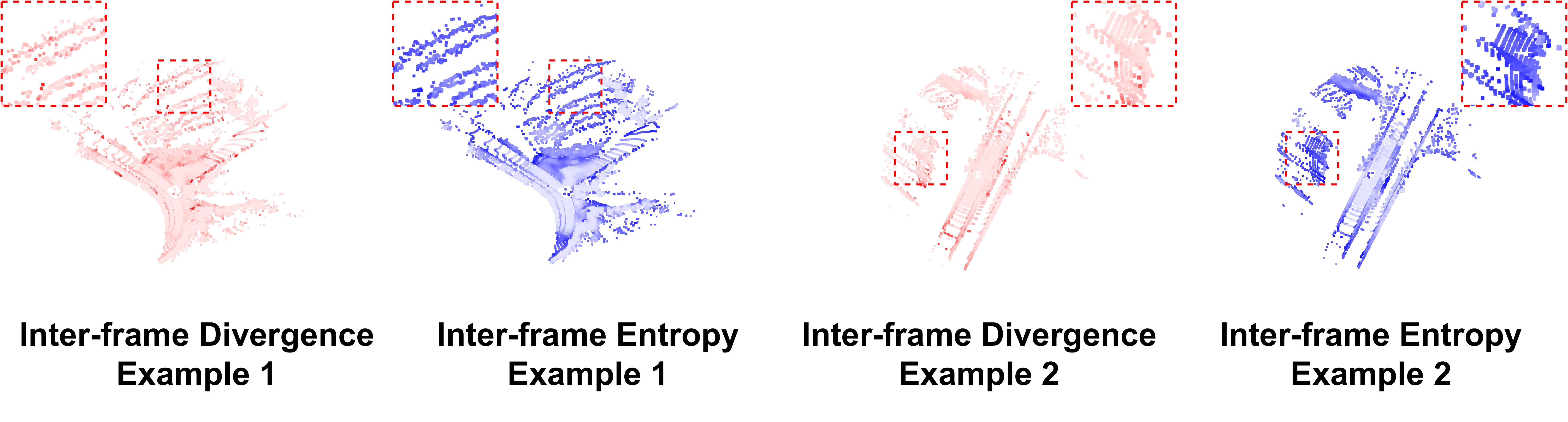}
\caption{\textbf{Examples of uncertainty scores.} 
{Red and blue indicate inter-frame divergence and entropy, respectively. The darker the color, the higher the value.}
Due to the sparsity and varying-density property of LiDAR point clouds, neural networks tend to generate class distributions that are more uniform for farther away sparse points. As highlighted by the dotted red boxes, this property results in misleadingly high values {for} 
far away sparse points in terms of inter-frame entropy but affect{s} less on the inter-frame divergence scores. 
}
\label{fig:example}
\end{figure}

Since the inter-frame divergence indicates {that} for each 
{region}
how inconsistent the predictions are across frames, the scores are similar for all the 
{regions}
that are in correspondence. To determine which one of the regions in correspondence contains the largest amount of beneficial information to improve the network, we retrieve the set of regions representing the same part of the outdoor scene and denote this set as $S$. We then look for the region from $S$ with 
the highest inter-frame entropy score:
    \begin{equation}
    (k, t) = \mathop{\arg\max}_{(j, s) \in S}\{{FE}_j^s | (j, s) \text{ and } (i, r) \text{ overlap}\},
    \end{equation}
where $(k, t)$ refers to the selected region $t$ from frame $F_k$. {Implementation details can be found in \textbf{Supplementary Section A}.}

The selected region is added to $D_L$ and all regions in set $S$ are then removed from $D_U$
{to avoid label redundancy.} The process is repeated until reaching the labeling budget. The active learning algorithm is summarized in Alg.~\ref{alg:active}.

One possible alternative strategy is to first find the region with the highest inter-frame entropy score and then select the one with the highest inter-frame divergence score in the corresponding set. A similar strategy is implemented in a previous work operating on multi-view images~\cite{siddiqui2020viewal}. However, unlike 2D images with dense and uniformly sampled pixels, 3D LiDAR frames have sparse and varying-density points. Specifically, the farther from the scanning viewpoint, the sparser the LiDAR points. As shown in Fig.~\ref{fig:example}, due to the sparsity, neural networks tend to predict uniform class distributions for peripheral points. This property will {result} 
in misleadingly high inter-frame entropy scores for points far away from the scanning viewpoint (Equation~\ref{equ:entropy}), while the inter-frame divergence scores remain stable (Equation~\ref{equ:divergence}). Considering the robustness of the system, we opt for the proposed strategy instead of the possible alternative. A quantitative comparison can be found in Section~\ref{sec:Ablation}. 

\subsubsection{Self-training.}
To further exploit the inter-frame constraints embedded in the LiDAR sequences, we leverage the fact that our measure of viewpoint inconsistency can also help us identify reliable regions with high-quality pseudo-labels, which can be directly injected into the training set. 

On the contrary with respect to active learning, which selects samples with the most uncertain predictions, self-training aims at acquiring confident and accurate pseudo-labels. To this end, we conduct a reversed process of the proposed active learning strategy. Specifically, we first look for the region $r$ from frame $F_i$ that has the lowest inter-frame divergence score in $D_U$:
    \begin{equation}
    (i, r) = \mathop{\arg\min}_{(j, s) \in D_U}{FD}_j^s,
    \end{equation}
We then look for the region from the corresponding set $S$ that has the lowest inter-frame entropy score:
    \begin{equation}
    (k, t) = \mathop{\arg\min}_{(j, s) \in S}\{{FE}_j^s | (j, s) \text{ and } (i, r) \text{ overlap}\},
    \end{equation}

The pseudo-label of the selected region is retrieved from the network predictions and all regions in set $S$ will not be used for further pseudo-labeling. The process is repeated until reaching the target 
number
{of pseudo-label{s}}. In order to prevent label drifting~\cite{feng2021active}, we reset the pseudo label set at each iteration and only select regions that are not already selected in the previous iteration. The self-training algorithm is summarized in Alg.~\ref{alg:self}.

\begin{minipage}{0.46\textwidth}
\begin{algorithm}[H]
    \centering
    \caption{Active Learning}\label{alg:active}
    \footnotesize
    \begin{algorithmic}
    \Require 
    \State $\text{Data set }D, \text{labeled set }D_L,$
    \State $\text{annotation budget } B, \text{metric } M$
    \Ensure 
    \State $\text{Added samples } A \gets \{\}$
    \State $\text{Unlabeled set } D_U \gets D \setminus D_L$
    \Repeat
        \State $(i, r) \gets \mathop{\arg\max}_{(j, s) \in D_U}{M_{FD}}_j^s$
        \State \text{Retrieve }$S$ \Comment{Corresponding set}
        \State $(k, t) \gets \mathop{\arg\max}_{(j, s) \in S}{M_{FE}}_j^s$
        \State $ A \gets A \cup (k, t)$
        \State $D_L \gets D_L \cup (k, t)$
        \State $D_U \gets D_U \setminus S$
    \Until $|A| = B \text{ or } |D_U| = 0$
    \State \textbf{return } $A$
    \end{algorithmic}
\end{algorithm}
\end{minipage}
\hfill
\begin{minipage}{0.46\textwidth}
\begin{algorithm}[H]
    \centering
    \caption{Self-training}\label{alg:self}
    \footnotesize
    \begin{algorithmic}
    \Require 
    \State $\text{Data set }D, \text{labeled set }D_L,$
    \State $\text{previous pseudo set } P, $
    \State $\text{target number } T, \text{metric } M$
    \Ensure 
    \State $\text{New pseudo set } P' \gets \{\}$
    \State $D_U' \gets (D \setminus D_L) \setminus P$ \Comment{No re-labeling}
    \Repeat
        \State $(i, r) \gets \mathop{\arg\min}_{(j, s) \in {D_U'}}{M_{FD}}_j^s$
        \State \text{Retrieve }$S$ \Comment{Corresponding set}
        \State $(k, t) \gets \mathop{\arg\min}_{(j, s) \in S}{M_{FE}}_j^s$
        \State $ P' \gets P' \cup (k, t)$
        \State $D_U' \gets D_U' \setminus S$
    \Until $|P'| = T \text{ or } |D_U'| = 0$ 
    \State \textbf{return } $P'$
    \end{algorithmic}
\end{algorithm}
\end{minipage}

\section{Experiments}
To demonstrate the effectiveness of our proposed method, we now present various experiments conducted on two large-scale 3D LiDAR semantic segmentation datasets, i.e., SemanticKITTI~\cite{behley2019semantickitti} and nuScenes~\cite{caesar2020nuscenes}. We first introduce the datasets and evaluation metrics 
in Section~\ref{sec:Datasets}, and then present the experimental settings in Section~\ref{sec:Exp_setting}. We report the results on the SemanticKITTI and nuScenes datasets in Section~\ref{sec:Results}, and the ablation studies in Section~\ref{sec:Ablation}.

\subsection{Datasets and Metrics} \label{sec:Datasets}

\noindent\textbf{SemanticKITTI~\cite{behley2019semantickitti}.} 
SemanticKITTI is a large-scale driving-scene dataset derived from the KITTI Vision Odometry Benchmark and {was} collected in Germany with the Velodyne-HDLE64 LiDAR. The dataset consists of 22 sequences containing 43{,}552 point cloud scans. We perform all our experiments using the official training (seq \text{00-07} and \text{09-10}) and validation (seq 08) split. 19 classes are used for segmentation. 

\noindent\textbf{nuScenes~\cite{caesar2020nuscenes}.}
nuScenes {was} 
collected in Boston and Singapore with 32-beam LiDAR sensors. It contains 1,000 scenes of 20s duration annotated with 2Hz frequency. Following the official train/val splits, we perform all label acquisition strategies on the 700 training sequences (28k scans) and evaluate them on 150 validation sequences (6k scans). 16 classes are used for segmentation.

\noindent\textbf{Metrics.}
For evaluation, we report mean class intersection over union (mIoU) results for both the SemanticKITTI and nuScenes datasets following the official guidance. 

\subsection{Experimental Settings} \label{sec:Exp_setting}

\subsubsection{Network Architectures.}
{To verify the effectiveness of the proposed active learning strategy on various network architectures,}
we {adopt} 
MinkowskiNet~\cite{choy20194d} based on sparse convolution, and SPVCNN~\cite{tang2020searching} based on point-voxel CNN, as our backbone networks for their great performance and high efficiency. We make the same choices {for {the} network architectures}
as ReDAL~\cite{wu2021redal}, a recent state-of-the-art active learning method of 3D semantic segmentation, for better comparison.

\subsubsection{Baseline Active Learning Methods.} 
{We select eight baseline methods for comparison, including} 
random frame selection ({RAND\textsubscript{fr}}), random region selection ({RAND\textsubscript{re}}), segment-entropy ({SEGENT})~\cite{lin2020efficient}, softmax margin ({MAR})~\cite{joshi2009multi,roth2006margin,wang2016cost}, softmax confidence ({CONF})~\cite{settles2008analysis,wang2016cost}, softmax entropy ({ENT})~\cite{hwa2004sample,wang2016cost},  core-set selection ({CSET})~\cite{sener2017active}, and {ReDAL}~\cite{wu2021redal}. The implementation details for all the methods are explained in \textbf{Supplementary Section B}.

\subsubsection{Learning Protocol.}
Following the same protocol as ReDAL~\cite{wu2021redal}, the model is initialized by training on $x_{init}\%$ of randomly selected LiDAR frames {with full annotations}.
The active learning process consists of $K$ rounds of the following actions: 1. Finetune the model on the current labeled set $D_L$. 2. Select $x_{active}\%$ of
data from the current unlabeled set $D_U$ for human annotation according to different active selection strategies. 3. Update $D_L$ and $D_U$.  

The labeling budget is measured by the {percentage} of labeled points. For both SemanticKITTI and nuScenes datasets, {we use} $x_{init} = 1\%$, $K = 4$, and $x_{active} = 1\%$. For self-training, the target 
number
of pseudo-labels {in terms of percentage of labeled points} $T = 1\%$. To ensure the reliability of the results, all the experiments are performed three times and the average results are reported. More training details can be found in \textbf{Supplementary Section A}.

\begin{figure}[t]
\centering
\includegraphics[width=\textwidth]{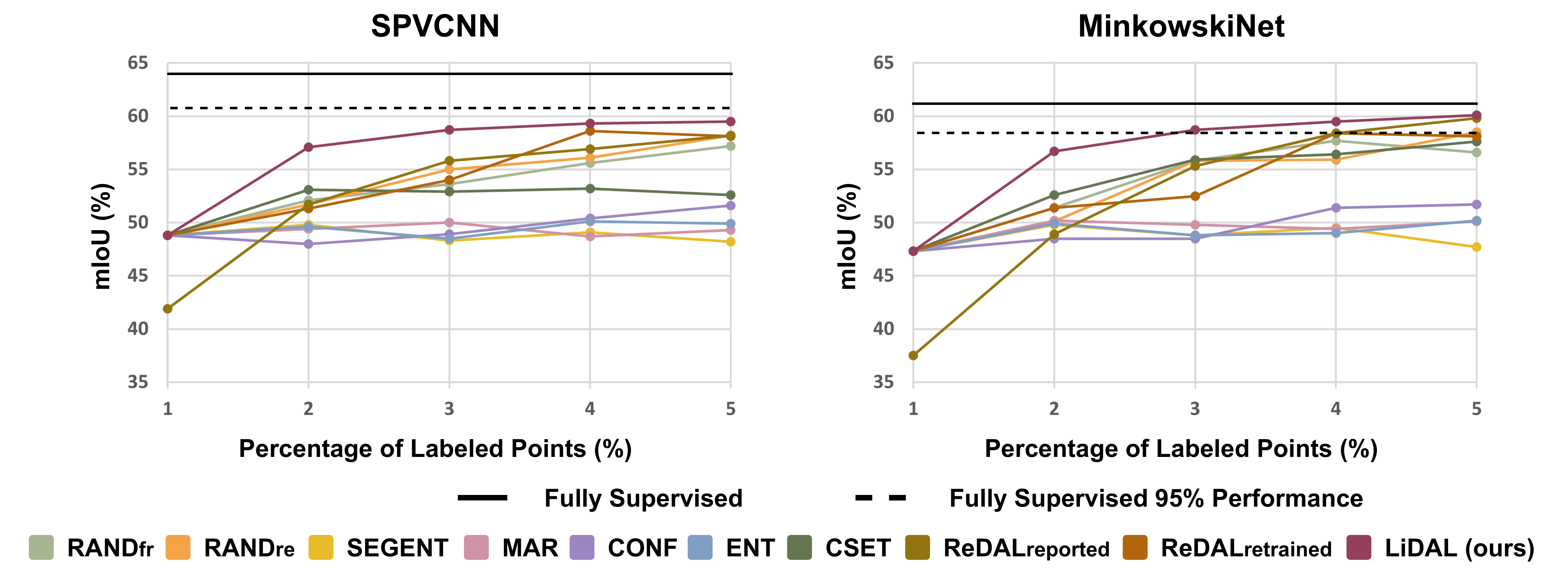}
\caption{\textbf{Mean intersection over union scores on SemanticKITTI Val~\cite{behley2019semantickitti}.} Detailed results can be found in \textbf{Supplementary Section C}.}
\label{fig:semantic}
\end{figure}

\subsection{Results and Analysis} \label{sec:Results}
\begin{figure}[t]
\centering
\includegraphics[width=\textwidth]{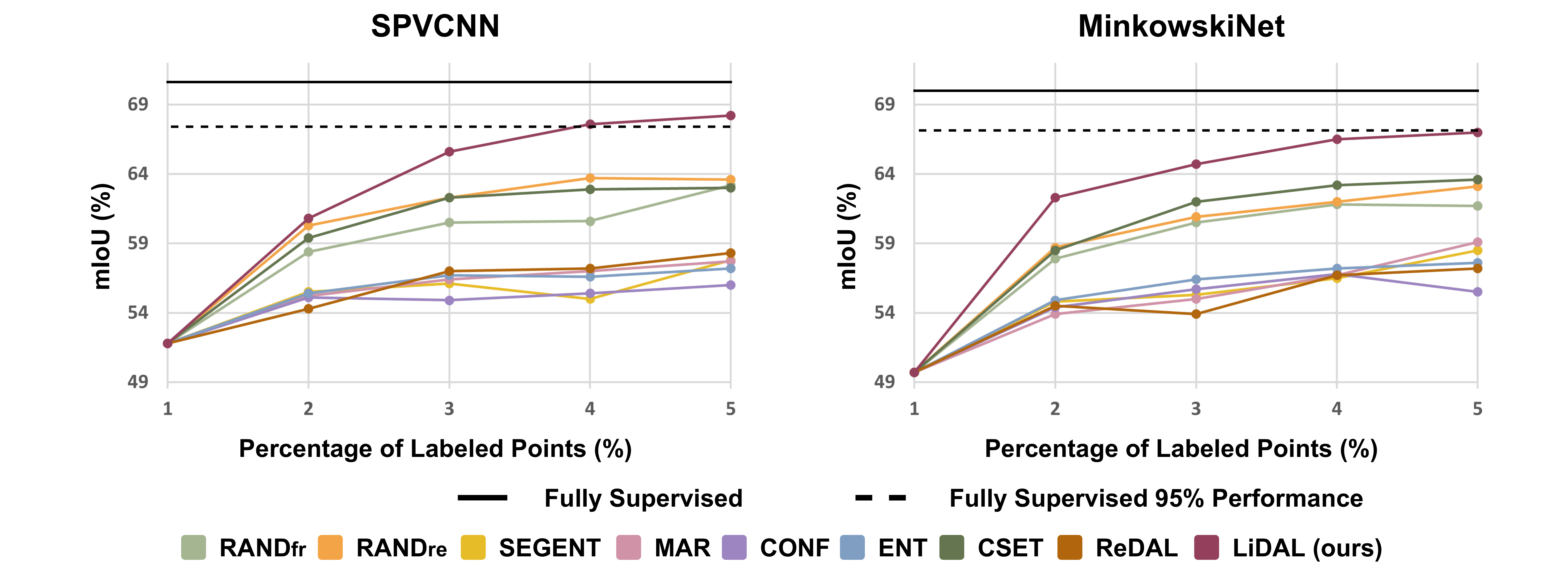}
\caption{\textbf{Mean intersection over union scores on nuScenes Val~\cite{caesar2020nuscenes}.} Detailed results can be found in \textbf{Supplementary Section C}.}
\label{fig:nuscenes}
\end{figure}

In this section, we present the performance of our approach compared to {the baseline methods} 
on the SemanticKITTI and nuScenes datasets. 
Fig.~\ref{fig:semantic} and Fig.~\ref{fig:nuscenes} {show the comparative results}. In each subplot, the x-axis represents the percentage of labeled points and the y-axis indicates the mIoU score achieved by the respective networks, which are trained with data selected through different active learning strategies. 

Since most {of the} baseline methods are not designed for LiDAR point clouds, 
{we re-implement these methods for LiDAR data based on their official codes.}
For {ReDAL}~\cite{wu2021redal}, in its published paper, it is evaluated on the SemanticKITTI dataset but not on the nuScenes dataset. For {the} SemanticKITTI dataset, we find that the reported scores of its initial networks (trained with 1\% of randomly selected frames) are way lower than our implementations (41.9 vs 48.8 for SPVCNN and 37.5 vs 47.3 for MinkowskiNet). 
We retrained its networks and {got} 
better results using its official code but with a finer training schedule (details can be found in \textbf{Supplementary Section A}). Both the retrained results and the reported results are presented in Fig.~\ref{fig:semantic}. For nuScenes, we adapt its official code and report the results.

\subsubsection{SemanticKITTI.}
As shown in Fig.~\ref{fig:semantic}, our proposed LiDAL significantly surpasses {the} 
existing active learning strategies 
{using}
the same percentage{s} of labeled data. Specifically, our method is able to reach nearly 95\% of the fully supervised performance with only 5\% of labeled data for the SPVCNN network and even achieve about 98\% of the fully supervised result for the MinkowskiNet network.   

In addition, we notice that many active learning strategies perform worse than the random baseline and some even bring negative effects on the network performance {(e.g., the mIoU scores may drop after adding the data samples selected by {SEGENT}).}
For uncertainty-based methods, such as {SEGENT} and {CONF}, since the model uncertainty values are calculated only within each frame 
{and are} 
biased by the peripheral points due to the scanning property of LiDAR, their performances are degraded. 
Even for pure diversity-based approaches, such as {CSET}, since the LiDAR datasets are captured as continuous frames and have plenty of redundant information among neighboring frames, simply clustering features may fail to produce diverse label acquisition.  

Moreover, we observe that the performance gap between {RAND\textsubscript{re}} and {RAND\textsubscript{fr}} is trivial. It showcases that if not combined with effective uncertainty measures, region-based training brings little benefit to the network performance.

\subsubsection{nuScenes.}
We also evaluate our algorithm on the novel nuScenes dataset and report the results in Fig.~\ref{fig:nuscenes}. As shown in the figure, our method outperforms all {the competitors} 
in terms of mIoU under all {the} experimental settings. Specifically, for both SPVCNN and MinkowskiNet, our method achieves more than 95\% of the fully supervised performances with only 5\% of labeled data.

\begin{table}
\small
\begin{center}
\caption{\textbf{Ablation study: building components.} FD: inter-frame divergence score; NMS: non-maximum suppression, i.e., select the region with the highest score in the corresponding set; FE: inter-frame entropy score.
}
\label{table::component}

\begin{tabular}{ |C{1.4cm}|C{2cm}|C{2cm}|C{1.4cm}|C{1.4cm}|C{1.4cm}|C{1.4cm}|}
    \hline
    FD & Frame-level & Region-level & NMS & FE & Pseudo & mIoU(\%) \\
    \hline
    \checkmark & \checkmark & & & & & 51.8 \\
    \hline
    \checkmark &  & \checkmark & & & & 52.5 \\
    \hline
    \checkmark &  & \checkmark & \checkmark & & & 55.5 \\
    \hline
    \checkmark &  & \checkmark & & \checkmark & & 56.4 \\
    \hline
    \checkmark &  & \checkmark & & \checkmark & \checkmark & 57.1 \\
    \hline
    \end{tabular}
\end{center}
\end{table}

Compared to the results of SemanticKITTI, similar phenomena can be witnessed that many active learning strategies perform worse than the random baseline. However, the negative effects are alleviated {and} {the mIoU scores 
consistently increase by adding data samples selected by most strategies}. A possible explanation is that, since the nuScenes dataset contains 1,000 scenes of tens of frames while the SemanticKITTI dataset contains only 22 scenes of thousands of frames, the network is less likely to be biased by {the} data selected from nuScenes than {that} from SemanticKITTI.

\subsection{Ablation Study} \label{sec:Ablation}
In this section, we conduct a number of controlled experiments that demonstrate the effectiveness of the building modules in LiDAL, and also examine some specific decisions in {our} LiDAL design. All the experiments are conducted on the SemanticKITTI validation set evaluating the performance of the SPVCNN network trained on the data selected in the first active learning round, keeping all {the} hyper-parameters the same. More ablation studies can be found in \textbf{Supplementary Section D}.

\subsubsection{Building Component{s}.}
In Table~\ref{table::component}, we evaluate the effectiveness of each component of our method.
\textbf{1. Effect of region-level labeling.}
``FD + Frame-level'' represents the baseline, which is to select frames with the highest average inter-frame divergence scores for training. By changing from ``FD + Frame-level'' to ``FD + Region-level'' (selecting regions), we can improve the performance by 0.7\%. This improvement is brought by focusing on the most informative parts of the scenes.
\textbf{2. Effect of active selection strategy.}
``FD + Region-level + NMS'' refers to selecting only the region with the highest inter-frame divergence score in the corresponding set. By avoiding the label redundancy, we can gain about 3\% of improvement. ``FD + Region-level + FE'' refers to the proposed selection strategy described in Section~\ref{sec:region}. From the proposed inter-frame entropy measure, we further improve about 0.9\%.
\textbf{3. Effect of pseudo-labels.}
``FD + Region-level + FE + Pseudo'' denotes the complete strategy of LiDAL. The introduction of pseudo-labels brings around 0.7\% of performance improvement. 

\begin{table}
\begin{center}
\caption{\textbf{Ablation study}: \textbf{(Left)} Region selection strategy; \textbf{(Right)} Target number of pseudo-label.}
\label{table::strategy}
    \begin{tabular}[b]{|C{3cm}|C{1.5cm}|}
    \hline
    Strategy & mIoU(\%) \\
    \hline
    FE + FD & 55.7 \\
    \hline
    FD + FE & 57.1 \\
    \hline
    \end{tabular}
    \quad
    \begin{tabular}[b]{|C{3cm}|C{1.5cm}|}
    \hline
    Target Number(\%) & mIoU(\%) \\
    \hline
    0.0 & 56.4\\
    \hline
    0.5 & 56.8\\
    \hline
    1.0 & 57.1\\
    \hline
    2.0 & 56.4\\
    \hline
    \end{tabular}
\end{center}
\end{table}

\subsubsection{Region Selection {Strategies}.
} 
In Section~\ref{sec:region}, we discuss two possible region selection strategies for both active learning and self-training. We advocate the proposed one that first finds corresponding sets using {the} inter-frame divergence scores and then selects regions with the inter-frame entropy scores. To justify our choice, we implement both strategies and report the results in Table~\ref{table::strategy} (Left). ``FD + FE'' refers to the proposed strategy and ``FE + FD'' refers to the possible alternative strategy that first finds corresponding sets using {the} entropy scores and then selects regions with {the} divergence scores. As shown in the table, the proposed strategy significantly outperforms the possible alternative strategy. It may be caused by the misleadingly high entropy values of peripheral points{, as} illustrated in Fig.~\ref{fig:example}.

\subsubsection{Target Number
of Pseudo-label{s}.}
In Section~\ref{sec:region}, we explore self-training in the proposed active learning framework and show that further gains can be realized by including pseudo-labels in Table~\ref{table::component}. To investigate the impact of pseudo-labels, we inject different
numbers
of pseudo-labels into the training set and report the results in Table~\ref{table::strategy} (Right). We observe that with the increasing
number
of pseudo-labels, the {gain of} network performance 
first increase{s} and then decrease{s}. 
{We speculate that adding pseudo-labels with a reasonable
number
will improve the network performance but superfluous pseudo-labels may bring unhelpful training biases and label noises. A further study on pseudo-labels can be found in \textbf{Supplementary Section D}.}

\section{Conclusion}
In this paper, we have presented a novel active learning strategy for 3D LiDAR semantic segmentation, named LiDAL. Aiming at exploiting the inter-frame constraints embedded in LiDAR sequences, we propose two uncertainty measures estimating the inconsistencies of network predictions among frames. We design a unified framework of both active learning and self-training {by} utilizing the proposed measures. Extensive experiments show that LiDAL achieves state-of-the-art results on the challenging SemanticKITTI and nuScenes datasets, significantly improving over strong baselines. For future works, one straightforward direction is to explore the potential of inter-frame constraints for RGB-D sequences of indoor scenes. Moreover, we believe that future works with special treatments for moving objects will further improve the performance.

\noindent\textbf{Acknowledgements.} This work is supported by Hong Kong RGC GRF 16206722 and a grant from City University of Hong Kong (Project No. 7005729).

%
%
\bibliographystyle{splncs04}
\bibliography{egbib}

\clearpage

\renewcommand\thesection{\Alph{section}}

\pagestyle{headings}
\mainmatter
\def\ECCVSubNumber{893}  

\title{
\emph{Supplementary Material} for \\
LiDAL: Inter-frame Uncertainty Based Active Learning for 3D LiDAR Semantic Segmentation
} 

\titlerunning{ECCV-22 submission ID \ECCVSubNumber} 
\authorrunning{ECCV-22 submission ID \ECCVSubNumber} 
\author{Anonymous ECCV submission}
\institute{Paper ID \ECCVSubNumber}

\maketitle

\begin{abstract}
This supplementary document is organized as follows:

\begin{itemize}
  \item Section \ref{sec:implementation} explains in more detail about the LiDAL implementation.
  \item Section \ref{sec:baseline} describes the baseline active learning methods.
  \item Section \ref{sec:experiment} enumerates detailed semantic segmentation results of the line charts in the main paper.
  \item Section \ref{sec:ablation} provides more ablation studies on {the} self-training strategy, class distribution of actively selected samples, and pseudo-label accuracy.

\end{itemize}
\end{abstract}

\section{Implementation Details
}\label{sec:implementation}
As explained in Section~3.1 of the main paper, our LiDAL method consists of four steps: 1. Train the network to convergence using the currently labeled dataset $D_L$. 2. Calculate the model uncertainty scores for each region of $D_U$. 3. Select regions based on the uncertainty measures for active learning and self-training. 4. Obtain labels from human annotation and pseudo-labeling. In this section, we supplement implementation details to these steps. Note that the used symbols are the same as those in Section~3 of the main paper.

\subsection{Network Training}
All the experiments are conducted on a PC with 8 NVIDIA Tesla V100 GPUs. The batch sizes are set to 30 and 90 for the SemanticKITTI~\cite{behley2019semantickitti} and nuScenes~\cite{caesar2020nuscenes} datasets, respectively.

For both datasets, we train the networks by minimizing the cross-entropy loss using Adam optimizer with an initial learning rate 1e-3. For fully-supervised baselines, the networks are trained for 80,000 iterations. For each round of active learning (including the initial round), the networks are trained or fine-tuned for 20,000 iterations.

The training settings are the same for SPVCNN~\cite{tang2020searching} and MinkowskiNet~\cite{choy20194d} network architectures.

\begin{figure}[t]
\centering
\includegraphics[width=\textwidth]{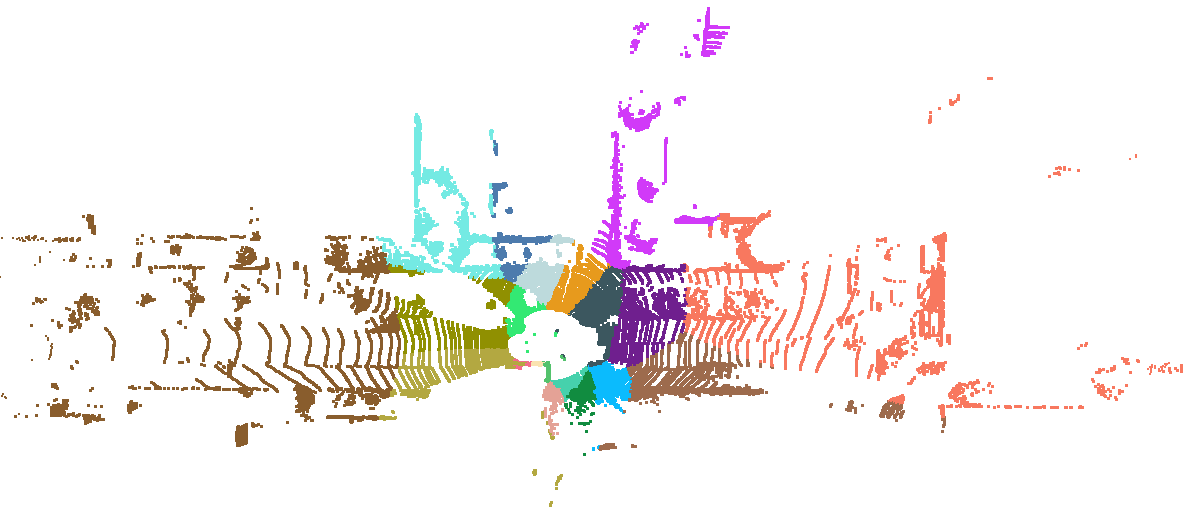}
\caption{\textbf{An example of divided sub-scene regions in the SemanticKITTI dataset.} Points of the same regions are painted with the same colors.
}
\label{fig:regions}
\end{figure}

\subsection{Correspondence Estimation}
In Section~3.2 of the main paper, after the registration of each frame, we then find for each point its corresponding points in {the} neighboring frames to calculate inter-frame uncertainty measures. Since there are hundreds of thousands of points in a LiDAR frame, it is impractical to register all the LiDAR frames at the same time and then estimate correspondences for each point.

To {address this issue}, 
for each frame {$F_i$}, we retrieve its neighboring $N_{nei}$ frames for correspondence estimation. After registration, for each point $p$ of the frame {$F_i$}, we find its nearest point in each of the neighboring $N_{nei}$ frames as the initial corresponding points. Since a certain position may be scanned in not all the frames due to occlusion and the movement of the scanning device, point $p$ may not have proper corresponding points in some neighboring frames. We then filter out the corresponding points {whose} 
distances to $p$ 
are larger than a threshold $T_{p}$. For both {the} SemanticKITTI and nuScenes datasets, we set $N_{nei} = 24$ and $T_{p} = 0.1m$.

\subsection{Region Division and Overlap Determining}
We utilize the constrained K-means clustering~\cite{bradley2000constrained} algorithm to divide a LiDAR frame $F$ into multiple sub-scene regions. As an extension of the classical K-means algorithm, this algorithm forces {the number of points in} each of the $K$ cluster{s} {in $(N_{min}, N_{max}$)}. 
For both {the} SemanticKITTI and nuScenes datasets, we set $K = 20$, $N_{min} = 0.95*\frac{|F|}{K}$, and $N_{max} = 1.05*\frac{|F|}{K}${, where} 
$|F|$ is the number of points contained in frame $F$.
An example of divided sub-scene regions of the SemanticKITTI dataset is shown in Fig~\ref{fig:regions}.

In Section~3.3 of the main paper, for a specific region $r$, we need to retrieve the set of regions overlapping with $r$ for further processing. To determine if two regions overlap, {we may check} 
the Earth Mover's distance~\cite{levina2001earth} or the Chamfer distance~\cite{butt1998optimum} between the two regions. However, we find that {a simpler solution based on} 
the distance between the weight centers of two regions yields similar results. 
Considering the efficiency {of this simple solution}, we determine {that} two regions overlap if the distance between their weight centers is less than $T_r$. For both the SemanticKITTI and nuScenes datasets, we set $T_r = 5m$.

\subsection{Label Acquisition}
For active learning, instead of using a {human} annotator, we simulate
annotation by using the ground-truth annotation of the dataset as the annotation from a 
human annotator. For self-training, we use the network predictions averaged over 8 augmented inference runs as the pseudo-labels.

\section{Baseline Active Learning Methods}\label{sec:baseline}
In this section, we describe the implementation of the baseline active learning methods used in our experiments (Section~4.2 of the main paper).

\subsubsection{Random Selection ({RAND\textsubscript{fr}} and {RAND\textsubscript{re}}).}
In each round of active learning, {this baseline method} randomly select{s} a portion of LiDAR frames or point cloud regions from the unlabeled dataset for label acquisition. It is a commonly used baseline strategy in the literature~\cite{wu2021redal,siddiqui2020viewal,sener2017active,feng2021active}.

\subsubsection{Segment-entropy (SEGENT).}
Based on the assumption that points within a region are supposed to share the same label, segment-entropy is proposed to serve as a metric for active selection~\cite{lin2020efficient}. In this method, the distribution of predicted labels within a region $r$ is estimated by:
    \begin{equation}
        E_{seg} = -\sum_c q(c) \log q(c),
    \end{equation}
    \begin{equation}
        \hat{y}^p = \mathop{\arg\max}_c P^p,        \end{equation}    
        \begin{equation}
        q(c) = \frac{1}{|r|}\sum_{p \in r} f(\hat{y}^p, c),    
    \end{equation}
        \begin{equation}
        f(\hat{y}^p, c) =
        \begin{cases}
            1,& \text{if } \hat{y}^p = c\\
            0,              & \text{otherwise}
        \end{cases},
    \end{equation}
where $E_{seg}$ is the proposed segment-entropy, $P^p$ is the probability distribution of point $p$, 
$\hat{y}^p$ is the predicted label of point $p$, {and} $q(c)$ is the percentage of points predicted as class $c$. The segment-entropy score of a frame is the average of the scores of all the points {inside this frame}. The frames with the largest segment-entropy scores are selected for label acquisition. In the implementation of this method, we utilize the same region division results as our LiDAL {for a fair comparison}. 

\subsubsection{Softmax Margin (MAR).}
Some previous active learning methods~\cite{joshi2009multi,roth2006margin,wang2016cost} rank all the samples in order of the model decision margin, which is the difference of softmax probabilities {between} 
the most probable label and the second most probable label, and then select the samples with the least differences. For a point $p$, the softmax margin is calculated as:
    \begin{equation}
        MAR^p = P^p(\hat{y}^1) - P^p(\hat{y}^2),
    \end{equation}
where $P^p$ is the probability distribution of point $p$, $\hat{y}^1$ is the {most} 
probable label class, and $\hat{y}^2$ is the second most probable label class.

The softmax margin of a frame is calculated by averaging the values of all the points {inside it}. The frames with the least softmax margin values are selected for label acquisition.

\subsubsection{Softmax Confidence (CONF).}
Similar to MAR, the softmax probability of the most probable label is considered as {a} confidence score in some previous methods~\cite{settles2008analysis,wang2016cost}. For point $p$, the softmax confidence is calculated as:
    \begin{equation}
        CONF^p = P^p(\hat{y}^1),
    \end{equation}
where $P^p$ is the probability distribution of point $p$, and $\hat{y}^1$ is the {most}
probable label class.

For a frame, the softmax confidence score is the average result of the scores of all the {associated} points. The frames with the least confidence scores are selected for label acquisition.

\subsubsection{Softmax Entropy (ENT).}
Unlike MAR and CONF, which consider only the top two most probable classes, softmax entropy takes into account probabilities of all classes to measure the information of a probability distribution~\cite{hwa2004sample,wang2016cost}. For point $p$, the softmax entropy score is calculated as:
    \begin{equation}
        ENT^p = -\sum_{c}P^p(c)\log (P^p(c)),
    \end{equation}
where $P^p(c)$ is the probability of point $p$ belonging to class $c$.

For a frame, the softmax entropy score is the average result of the scores of all the {associated} points. The frames with the largest entropy scores are selected for label acquisition.

\subsubsection{Core-set Selection (CSET).}

Core-set refers to a small subset that captures the diversity of the whole dataset~\cite{sener2017active}, {and} thus a model trained on {this subset} 
yields similar performance to that trained on the whole dataset. This method first extracts features for each sample of the dataset using the currently trained network. Operating on the feature space, it then selects a small set of samples for labeling utilizing the furthest point sampling strategy. In the implementation, we use the intermediate results of the second-last layers of the networks as the features. The feature of a frame is averaged over all the associated points.

\subsubsection{ReDAL.}

Region-based and diversity-aware active learning (ReDAL)~\cite{wu2021redal} is a recent state-of-the-art method designed for 3D semantic segmentation of both indoor and outdoor scenes. This method first divides a 3D scene into sub-scene regions and then estimates the region information utilizing three metrics: softmax entropy, color discontinuity, and structural complexity. With the estimated region information scores, this method further designs a diversity-aware selection algorithm to avoid visually similar regions appearing in a querying batch for labeling. Since both the SemanticKITTI and nuScenes datasets do not provide colored point clouds, the color discontinuity metric is discarded in the implementation following the instruction of ReDAL's official code.

\section{Detailed Experimental Results}\label{sec:experiment}

In this section, we provide more details on our experimental results, for benchmarking purposes with future works. The results of fully-supervised networks are reported in Table~\ref{table::full}. Detailed scores for
Fig.~5 in the main paper are shown in Tables~\ref{table::fig5_spvcnn} and~\ref{table::fig5_mink}. For Fig.~6, the detailed scores are presented in Tables~\ref{table::fig6_spvcnn} and~\ref{table::fig6_mink}. 

\begin{table}
\small
\begin{center}
\caption{\textbf{Mean intersection over union scores of fully-supervised networks.}
}
\label{table::full}

\begin{tabular}{|C{4cm}|C{3cm}|C{3cm}|}
    \hline
    Network $\setminus$ Dataset & SemanticKITTI & nuScenes \\
    \hline
    SPVCNN & 64.5 & 71.7 \\
    \hline
    MinkowskiNet & 61.4 & 70.6 \\
    \hline
    \end{tabular}
\end{center}
\end{table}

\begin{table}
\small
\begin{center}
\caption{\textbf{Mean intersection over union scores on SemanticKITTI Val with SPVCNN.}
}
\label{table::fig5_spvcnn}

\begin{tabular}{|C{4.5cm}|C{1.4cm}|C{1.4cm}|C{1.4cm}|C{1.4cm}|C{1.4cm}|}
    \hline
    Percentage of Labeled Points & Init (1\%) & 2\% & 3\% & 4\% & 5\% \\
    \hline
    RAND\textsubscript{fr} & 48.8 & 52.1 & 53.6 & 55.6 & 57.2 \\
    \hline
    RAND\textsubscript{re} & 48.8 & 51.7 & 55.0 & 56.1 & 58.2 \\
    \hline
    SEGENT & 48.8 & 49.8 & 48.3 & 49.1 & 48.2 \\
    \hline
    MAR & 48.8 & 49.4 & 50.0 & 48.7 & 49.3 \\
    \hline
    CONF & 48.8 & 48.0 & 48.9 & 50.4 & 51.6 \\
    \hline
    ENT & 48.8 & 49.6 & 48.5 & 50.1 & 49.9 \\
    \hline
    CSET & 48.8 & 53.1 & 52.9 & 53.2 & 52.6 \\
    \hline
    ReDAL\textsubscript{reported} & 41.9 & 51.7 & 55.8 & 56.9 & 58.2 \\
    \hline
    ReDAL\textsubscript{retrained} & 48.8 & 51.3 & 54.0 & 58.6 & 58.1 \\ 
    \hline
    LiDAL (ours) & 48.8 & 57.1 & 58.7 & 59.3 & 59.5 \\
    \hline
    \end{tabular}
\end{center}
\end{table}

\begin{table}
\small
\begin{center}
\caption{\textbf{Mean intersection over union scores on SemanticKITTI Val with MinkowskiNet.}
}
\label{table::fig5_mink}

\begin{tabular}{|C{4.5cm}|C{1.4cm}|C{1.4cm}|C{1.4cm}|C{1.4cm}|C{1.4cm}|}
    \hline
    Percentage of Labeled Points & Init (1\%) & 2\% & 3\% & 4\% & 5\% \\
    \hline
    RAND\textsubscript{fr} & 47.3 & 51.4 & 55.8 & 57.7 & 56.6 \\
    \hline
    RAND\textsubscript{re} & 47.3 & 50.1 & 55.8 & 55.9 & 58.5 \\
    \hline
    SEGENT & 47.3 & 49.8 & 48.8 & 49.5 & 47.7 \\
    \hline
    MAR & 47.3 & 50.2 & 49.8 & 49.4 & 50.1 \\
    \hline
    CONF & 47.3 & 48.5 & 48.5 & 51.4 & 51.7 \\
    \hline
    ENT & 47.3 & 49.9 & 48.8 & 49.0 & 50.2 \\
    \hline
    CSET & 47.3 & 52.6 & 55.9 & 56.4 & 57.6 \\
    \hline
    ReDAL\textsubscript{reported} & 37.5 & 48.9 & 55.3 & 58.4 & 59.8 \\
    \hline
    ReDAL\textsubscript{retrained} & 47.3 & 51.4 & 52.5 & 58.4 & 58.1 \\ 
    \hline
    LiDAL (ours) & 47.3 & 56.7 & 58.7 & 59.5 & 60.1 \\
    \hline
    \end{tabular}
\end{center}
\end{table}

\begin{table}
\small
\begin{center}
\caption{\textbf{Mean intersection over union scores on nuScenes Val with SPVCNN.}
}
\label{table::fig6_spvcnn}

\begin{tabular}{|C{4.5cm}|C{1.4cm}|C{1.4cm}|C{1.4cm}|C{1.4cm}|C{1.4cm}|}
    \hline
    Percentage of Labeled Points & Init (1\%) & 2\% & 3\% & 4\% & 5\% \\
    \hline
    RAND\textsubscript{fr} & 51.8 & 58.4 & 60.5 & 60.6 & 63.2 \\
    \hline
    RAND\textsubscript{re} & 51.8 & 60.3 & 62.3 & 63.7 & 63.6 \\
    \hline
    SEGENT & 51.8 & 55.5 & 56.1 & 55 & 57.8 \\
    \hline
    MAR & 51.8 & 55.2 & 56.4 & 57.0 & 57.7 \\
    \hline
    CONF & 51.8 & 55.1 & 54.9 & 55.4 & 56.0 \\
    \hline
    ENT & 51.8 & 55.4 & 56.7 & 56.6 & 57.2 \\
    \hline
    CSET & 51.8 & 59.4 & 62.3 & 62.9 & 63.0 \\
    \hline
    ReDAL & 51.8 & 54.3 & 57.0 & 57.2 & 58.3 \\
    \hline
    LiDAL (ours) & 51.8 & 60.8 & 65.6 & 67.6 & 68.2 \\
    \hline
    \end{tabular}
\end{center}
\end{table}

\begin{table}
\small
\begin{center}
\caption{\textbf{Mean intersection over union scores on nuScenes Val with MinkowskiNet.}
}
\label{table::fig6_mink}

\begin{tabular}{|C{4.5cm}|C{1.4cm}|C{1.4cm}|C{1.4cm}|C{1.4cm}|C{1.4cm}|}
    \hline
    Percentage of Labeled Points & Init (1\%) & 2\% & 3\% & 4\% & 5\% \\
    \hline
    RAND\textsubscript{fr} & 49.7 & 57.9 & 60.5 & 61.8 & 61.7 \\
    \hline
    RAND\textsubscript{re} & 49.7 & 58.7 & 60.9 & 62.0 & 63.1 \\
    \hline
    SEGENT & 49.7 & 54.8 & 55.3 & 56.5 & 58.5 \\
    \hline
    MAR & 49.7 & 53.9 & 55.0 & 56.7 & 59.1 \\
    \hline
    CONF & 49.7 & 54.4 & 55.7 & 56.8 & 55.5 \\
    \hline
    ENT & 49.7 & 54.9 & 56.4 & 57.2 & 57.6 \\
    \hline
    CSET & 49.7 & 58.5 & 62.0 & 63.2 & 63.6 \\
    \hline
    ReDAL &49.7 & 54.5 & 53.9 & 56.7 & 57.2 \\
    \hline
    LiDAL (ours) & 49.7 & 62.3 & 64.7 & 66.5 & 67.0 \\
    \hline
    \end{tabular}
\end{center}
\end{table}

\section{More Ablation Studies}\label{sec:ablation}

In this section, we provide more ablation studies to examine the design decision of our self-training strategy and to analyse the actively selected labels and pseudo-labels.

\subsubsection{Self-training Strategy.}
In Section~3.3 of the main paper, we inject pseudo-labels to the training set at each active learning round to further boost the performance. We have considered three commonly used strategies for self-training:

\begin{itemize}
    \item \textbf{S1:} Enlarge the pseudo-label set in each round with the newly selected regions. (The selection criterion is discussed in the main paper.)
    \item \textbf{S2:} Keep the size of the pseudo-label set constant, and replace in each round with the newly selected regions.
    \item \textbf{S3:} (Our design choice) Keep the size of the pseudo-label set constant, and replace in each round with the newly selected regions that are not already in the last pseudo-label set.
\end{itemize}

The results of these three self-training strategies on the SemanticKITTI dataset with SPVCNN are shown in Table~\ref{table::st_strategy}. As shown in the table, both two alternative strategies generate more inferior results {to} 
our design choice. We assume that, for \textbf{S1}, it is easily susceptible to label drifting as its size of pseudo-label set increases over time. {For} 
\textbf{S2}, since the previous pseudo-label set used for training is also considered for the pseudo-labeling of the current round, it tends to select a stable set of regions that are less and less helpful during training.

\begin{table}
\small
\begin{center}
\caption{\textbf{Mean intersection over union scores of different self-training strategies on SemanticKITTI Val with SPVCNN.}
}
\label{table::st_strategy}

\begin{tabular}{|C{4.5cm}|C{1.4cm}|C{1.4cm}|C{1.4cm}|C{1.4cm}|C{1.4cm}|}
    \hline
    Percentage of Labeled Points & Init (1\%) & 2\% & 3\% & 4\% & 5\% \\
    \hline
    \textbf{S1} & 48.8 & 56.9 & 57.2 & 59.1 & 58.8 \\
    \hline
    \textbf{S2} & 48.8 & 57.2 & 58.5 & 58.9 & 59.0 \\
    \hline
    \textbf{S3} (Our design choice) & 48.8 & 57.1 & 58.7 & 59.3 & 59.5 \\
    \hline
    \end{tabular}
\end{center}
\end{table}

\subsubsection{Class Distribution of Actively Selected Samples.}
To gain a better understanding of the property of inter-frame constraints, we count the class distribution of samples selected in all 4 rounds of LiDAL operating on the SemanticKITTI dataset with SPVCNN network. As shown in Table~\ref{table:class}, LiDAL focuses more on less-represented but highly important classes like person and bicyclist. This is foreseeable since {the} networks struggle to generate consistent predictions for these hard samples. 
This is a valuable property that can benefit downstream tasks like autonomous driving, which poses great significance on safety issues.

\begin{table}
\begin{center}
\caption{\textbf{Class distributions of labels(\textperthousand).} We present samples selected in all 4 rounds of LiDAL operating on the SemanticKITTI dataset with SPVCNN network.
}
\label{table:class}
\resizebox{\textwidth}{!}{
    \begin{tabular}{  | l | c | c c c c c c c c c c c c c c c c c c c|}
    \hline
    Method & Total & \rotatebox[origin=c]{90}{car} & \rotatebox[origin=c]{90}{bicycle} & \rotatebox[origin=c]{90}{motorcycle} & \rotatebox[origin=c]{90}{truck} & \rotatebox[origin=c]{90}{other-vehicle} & \rotatebox[origin=c]{90}{person} & \rotatebox[origin=c]{90}{bicyclist} & \rotatebox[origin=c]{90}{motorcyclist} & \rotatebox[origin=c]{90}{road} & \rotatebox[origin=c]{90}{parking} & \rotatebox[origin=c]{90}{sidewalk} & \rotatebox[origin=c]{90}{other-ground} & \rotatebox[origin=c]{90}{building} & \rotatebox[origin=c]{90}{fence} & \rotatebox[origin=c]{90}{vegetation} & \rotatebox[origin=c]{90}{trunk} & \rotatebox[origin=c]{90}{terrain} & \rotatebox[origin=c]{90}{pole} & \rotatebox[origin=c]{90}{traffic-sign}\\
    \hline
    Full & $10^3$ & 43.68 & 0.17 & 0.42 & 2.02 & 2.40 & 0.36 & 0.13 & 0.04 & 205.22 & 15.19 & 148.59 & 4.03 & 137.00 & 74.69 & 275.57 & 6.23 & 80.67 & 2.95 & 0.63 \\
    \hline
    LiDAL & $10^3$ & 36.75 & 0.25 & 1.06 & 4.01 & 7.45 & 0.89 & 0.39 & 0.07 & 146.42 & 22.30 & 154.42 & 11.91 & 127.15 & 98.29 & 277.80 & 9.01 & 95.91 & 4.34 & 1.57 \\
    \hline
    \end{tabular}}
\end{center}
\end{table}

\subsubsection{Accuracy of Pseudo-labels.}
The main challenge with pseudo-labels is to ensure their accuracy and to avoid drifting. In Section~4.4 of the main paper, we evaluate the effects of injecting different numbers of pseudo-labels into the training set. Here we quantitatively measure the accuracy of added pseudo-labels in Table~\ref{table::accuracy}. The study is conducted {in} 
the first training round of SPVCNN on the SemanticKITTI dataset. As shown in the table, the generated pseudo-labels maintain high accuracy in general, but the accuracy drops when more and more pseudo-labels are selected. This confirms our conjecture in the main paper that adding a reasonable number of pseudo-labels will improve network performance, but redundant pseudo-labels might introduce unhelpful training bias and label noise.

\begin{table}
\small
\begin{center}
\caption{\textbf{Accuracy of pseudo-labels.} Samples are selected in the first training round of SPVCNN on SemanticKITTI dataset. 
}
\label{table::accuracy}

\begin{tabular}{|C{6cm}|C{4cm}|}
    \hline
    Range of Added Pseudo-labels & Mean Accuracy \\
    \hline
    0-1\% & 97.58\% \\
    \hline
    1-2\% & 97.04\% \\
    \hline
    2-3\% & 93.05\% \\
    \hline
    \end{tabular}
\end{center}
\end{table}

%
%

\end{document}